%% file: main.tex
\documentclass[letterpaper, 10 pt, conference]{ieeeconf}  

\IEEEoverridecommandlockouts                              

\usepackage{graphicx}
\usepackage{amsmath, amsfonts, amssymb} 
\usepackage{algorithmic, algorithm}
\usepackage{acro}
\usepackage{bm}
\input{include/acronyms.tex}
\input{include/newcommands.tex}
\usepackage{siunitx}
\usepackage{subcaption}
\usepackage{booktabs}
\usepackage{multirow}
\usepackage{placeins}
\usepackage{cite}
\usepackage{hyperref}
\usepackage{caption}
\newif\ifshowtable

\newif\ifshowfigure
\showfiguretrue

\title{\LARGE \bf
Energy-Optimized Planning in Non-Uniform Wind Fields\\with Fixed-Wing Aerial Vehicles
}

\author{Yufei Duan*$^{,1,2}$, Florian Achermann*$^{,2}$, Jaeyoung Lim$^{2}$, Roland Siegwart$^{2}$
\thanks{This work was supported by ETH Research Grant AvalMapper ETH-10 20-1 and Knut and Alice Wallenberg Foundation.
}
\thanks{* These authors contributed equally to this work}%
\thanks{$^{1}$ Robotics, Perception and Learning Lab, KTH Royal Institude of Technology, Stockholm 114 28, {\tt \footnotesize \{yufeidu\}@kth.se}}%
\thanks{$^{2}$ Autonomous Systems Lab, ETH Z\"urich, Z\"urich 8092, Switzerland {\tt \footnotesize \{yuduan, acfloria, jalim, rsiegwart\}@ethz.ch}}%
}

\begin{document}

\maketitle
\thispagestyle{empty}
\pagestyle{empty}

\begin{abstract}

Fixed-wing \acp{sUAV} possess the capability to remain airborne for extended durations and traverse vast distances. However, their operation is susceptible to wind conditions, particularly in regions of complex terrain where high wind speeds may push the aircraft beyond its operational limits, potentially raising safety concerns. Moreover, wind impacts the energy required to follow a path, especially in locations where the wind direction and speed are not favorable. Incorporating wind information into mission planning is essential to ensure both safety and energy efficiency. In this paper, we propose a sampling-based planner using the kinematic Dubins aircraft paths with respect to the ground, to plan energy-efficient paths in non-uniform wind fields. We study the characteristics of the planner with synthetic and real-world wind data and compare its performance against baseline cost and path formulations. We demonstrate that the energy-optimized planner effectively utilizes updrafts to minimize energy consumption, albeit at the expense of increased travel time. The ground-relative path formulation facilitates the generation of safe trajectories onboard \acp{sUAV} within reasonable computational timeframes.

\end{abstract}

\section{INTRODUCTION}

Recent advances in \acp{sUAV} have enabled various tasks taken by \acp{sUAV} including mapping~\cite{COLOMINA2014unmanned}, inspection, and environmental monitoring~\cite{ezequiel2014uav}. Fixed-wing type \acp{sUAV} have been popular, especially for large-scale environmental monitoring tasks due to their long endurance and capability to cover a large area. Though efficient, they are also sensitive to wind, which can significantly influence the performance of the vehicle~\cite{schopferer2015performance}, and even compromise safety~\cite{stastny2019flying}. This has resulted in the deployments of fixed-wing \acp{sUAV} limited only to large open spaces and moderate wind conditions.

Operating fixed-wing~\acp{sUAV} in diverse wind conditions close to the terrain will enable various applications that have otherwise been carried out by inefficient multirotor vehicles~\cite{boon2017comparison}. However, operating fixed-wing~\acp{sUAV} close to the terrain poses a significant challenge. The limited maneuverability and strong winds induced by the terrain impose a high risk of collision, as wind speeds can easily exceed the limitations of fixed-wing \acp{sUAV}~\cite{stastny2019flying, ch_windatlas}.

Previous studies addressing fixed-wing \ac{sUAV} operations close to the terrain have predominantly focused on planning scenarios without considering wind~\cite{lim2024safe} or uniform wind fields~\cite{schopferer2015performance, moon2023time, techy2009minimum, bucher2023robust}. However, the assumption that the wind is uniform is unrealistic for large-scale navigation, as the wind changes in proximity to the terrain and across different height levels. While planning in non-uniform wind fields has been considered for soaring~\cite{chung2015learning, chakrabarty2013uav}, global planning utilizing a discretized grid~ \cite{wirth2015}, and close to the ground for minimum time~\cite{oettershagen2017towards}, they are either computationally too expensive or ignoring the non-holonomic constraints of fixed-wing vehicles. Considering the non-uniform wind field together with the fixed-wing \ac{sUAV}'s kinematic limits allows for safe planning. The spatial variation in the wind field provides an opportunity to exploit the energy available in the atmosphere for more efficient flight, thus extending the endurance and range of the vehicle.


\begin{figure}[t]
\vspace{2mm}
\centerline{\includegraphics[width=\linewidth]{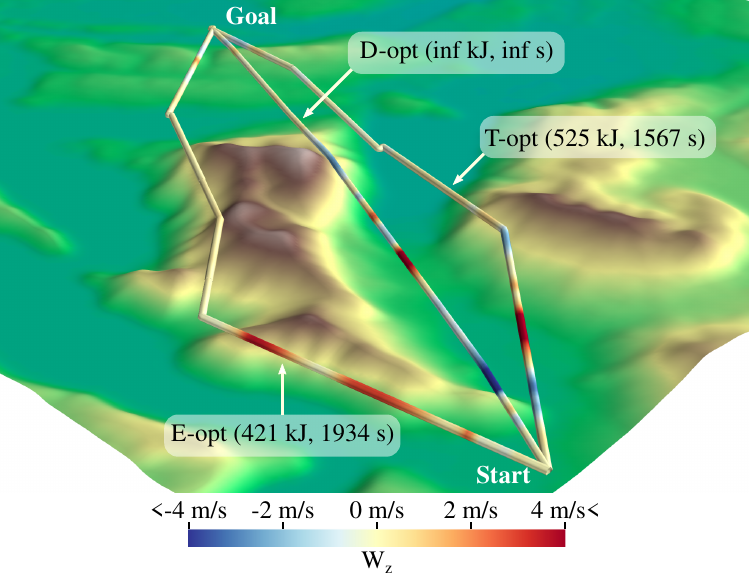}}
\caption{Planned paths around mountainous terrain with different cost formulations colored by the encountered vertical wind ($W_z$). The energy-optimized path (E-opt) leverages the updrafts on the left side of the mountain leading to a path with lower energy required but longer flight time compared to the time-optimized path (T-opt). The minimum-distance path (D-opt) disregards any wind information and is infeasible to track due to the strong downdrafts encountered along the path.}
\vspace{-2mm}
\label{fig:real_case1}
\end{figure}

In this work, we present a novel approach that utilizes geometric Dubins airplane paths~\cite{chitsaz2007time} and incorporates temporally-constant non-uniform wind fields for energy-optimized planning. We analyze the advantages of employing a ground-relative Dubins airplane path over the computationally complex air-relative Dubins airplane path formulation~\cite{oettershagen2017towards}. Our method demonstrates the capability to exploit wind efficiently, resulting in significant reductions of energy consumption, while safely and successfully reaching the goal. We evaluate the impacts of using energy-optimized planning first in synthetic wind fields and later on realistic wind fields around complex terrains generated by the \ac{WRF}~\cite{skamarock2019wrf}. Ultimately, our planner finds the energy-optimized path given a known wind field. This method enables the vehicle to minimize the energy expenditure in reaching the target position, thereby enhancing its mission duration capacity and range, while ensuring safety, particularly in adverse wind conditions.  In addition, using energy as the planning objective enables the operators to assess pre-flight whether the \ac{sUAV} has enough fuel/energy onboard to complete the mission under the considered wind conditions.

Our key contributions are as follows:
\begin{itemize}
    \item We propose a sampling-based planning framework for energy-optimized planning using the Dubins airplane model in non-uniform wind fields.
    \item We evaluate and compare the trade-offs between the ground-relative and air-relative Dubins airplane paths.
    \item We compare the merits and drawbacks of energy-optimized path planning with those of the shortest path and time-optimized planning.
\end{itemize}

\section{RELATED WORK} \label{related}

Practical implementations of sampling-based planning methods for fixed-wing aerial vehicles have been demonstrated in various environments, such as cluttered indoor~\cite{bry2015aggressive} and steep alpine environments~\cite{lim2024safe, oettershagen2017towards}.

One of the main challenges in considering wind is representing the non-holonomic constraints with wind-relative kinematics. \cite{bry2015aggressive} uses polynomials to consider the smooth dynamics of the system for planning and approximates the polynomials with Dubins curves~\cite{dubins1957curves}. Dubins airplane paths~\cite{chitsaz2007time}, which extend the Dubins curve into three-dimensional space with additional climb rate constraints are most commonly used~\cite{oettershagen2017towards, mclain2014implementing}. Other works have considered using smooth path representations such as Bezier curves~\cite{seemann2014rrt}, and splines~\cite{lee2014optimal}. However, these works do not consider wind and therefore can result in suboptimal or even infeasible paths when deployed in realistic environments.

Many works consider a spatially uniform wind, which leads to path representations such as trochoids~\cite{schopferer2015performance, moon2023time, techy2009minimum, bucher2023robust}. Trochoids take the drift induced by the wind into account if Dubins curves are tracked in the air-relative frame. However, these methods typically require iteratively solving the path which is ill-suited for sampling-based planners due to the computational cost. Moreover, the uniform wind field assumption holds poorly for large-scale navigation tasks, especially when operating close to the terrain or obstacles where the spatial wind gradient can be significant~\cite{achermann2024windseer, patrikar2020wind}.

Planning in non-uniform wind fields for autonomous soaring has been investigated with a local reinforcement agent~\cite{chung2015learning} or a non-optimal tree-search-based method~\cite{chakrabarty2013uav}. Previous work presented energy-optimal planning on a large-scale discretized grid in non-uniform wind~\cite{wirth2015}. However, this work is not suitable for navigation around complex terrain due to the coarse grid discretization and neglecting the \ac{sUAV}'s kinematic limits.

Our work is a continuation of~\cite{oettershagen2017towards} which applies sampling-based planning in a non-uniform wind field to plan time-optimized paths. However, the iterative nature of finding the air-relative path significantly slows down the convergence speed of the planner, rendering it unsuitable for onboard deployment. In this work, we show that the use of geometric Dubins airplane paths speeds up the planner even while considering the energy objective.

\section{PROBLEM FORMULATION} \label{prob}

Consider a state space $\mathcal{X}$, comprising the allowed space $\mathcal{X}^+$ and the forbidden space $\mathcal{X}^- =\mathcal{X}/\mathcal{X}^+$. A subset of permitted states are safe states, denoted $\mathcal{X}_{safe} \subset \mathcal{X}^+$, which are designated safe for the operation of~\acp{sUAV}. We also assume that we have a known temporally constant wind field $\mathcal{W}:\mathcal{X} \mapsto \mathbb{R}^3$. We consider the problem of finding the minimum energy path $\bm{\eta^*}:[0, S] \mapsto \mathcal{X}$ 
 between the initial state $\bm{x}_{start}$ and the goal state $\bm{x}_{goal}$ while operating safely in a known wind field $\mathcal{W}$ with the aircraft's ground speed ${V}$, where $S$ is the length of the path. The problem can be formally written as:


\begin{align}
     \bm{\eta}^*=& \arg \min_{\bm{\eta}} \int_{\bm{\eta}}^{} P(\bm{\eta}(s), \bm{W})\frac{1}{V(\bm{\eta}(s), \bm{W})} d\bm{s}\\
     s.t. &\quad \bm{\eta}(s) \in \mathcal{X}_{safe} &&\forall s \in [0, S]\nonumber\\
     &\frac{\partial \bm{\eta}}{\partial s} = f(\bm{\eta}(s))\quad &&\forall s \in [0, S]\nonumber\\
     &\bm{\eta}(0) = \bm{x}_{start},  \bm{\eta}(S) = \bm{x}_{goal}\nonumber
\end{align}

Here, $P$ denotes the power consumption of the \ac{sUAV}. The path is required to stay within the safe set $\mathcal{X}_{safe}$, and to satisfy the nonholonomic dynamic constraint of the vehicle, defined as $f(\cdot)$. 

\subsection{Dubins Airplane Model}
We consider a state space that represents the vehicle $\bm{x}=(x,y,z,\theta^\mathcal{G})$, where $x, y, z$ is the three-dimensional position in space and the $\theta^\mathcal{G}$ is the ground relative bearing of the vehicle. We use the Dubins airplane model~\cite{chitsaz2007time} to represent the non-holonomic constraint of fixed-wing vehicles. The constraints can be defined as differential constraints:

\begin{align}
    \frac{\partial \bm{\eta}}{\partial s} = f(\bm{\eta}(s)) = \begin{pmatrix}\cos(\gamma^\mathcal{G})\cos(\theta^\mathcal{G})\\
         \cos(\gamma^\mathcal{G})\sin(\theta^\mathcal{G})\\
         \sin(\gamma^\mathcal{G})\\
         cos(\gamma^\mathcal{G}) \kappa
         \end{pmatrix},
    \label{eq:dubins_airplane}
\end{align}
where $\gamma^\mathcal{G}$ is the flight path angle, and $\kappa$ is curvature. The maneuverability of the vehicle is limited by the flight path angle $\gamma^\mathcal{G} \in [\gamma^{\mathcal{G}}_{min}, \gamma^{\mathcal{G}}_{max}]$ and curvature $\kappa \in [-\kappa_{max}, \kappa_{max}]$. 

\begin{figure}[t]
\vspace{2mm}
\centerline{\includegraphics[width=0.65\linewidth]{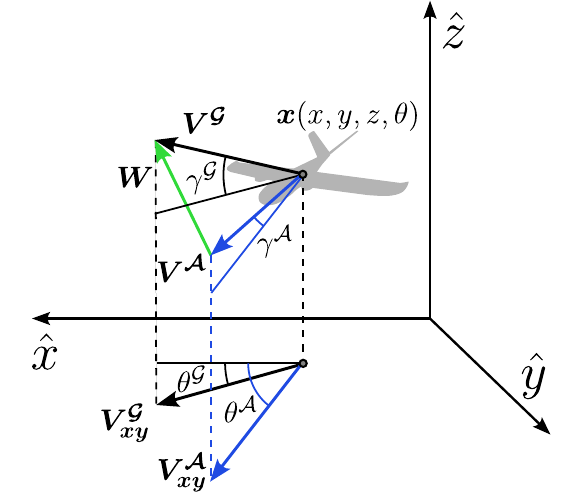}}
\caption{State space of the Dubins airplane model with the air relative (superscript $\mathcal{A}$) and ground-relative (superscript $\mathcal{G}$) properties.}
\label{fig:dubins_airplane}
\vspace{-1mm}
\end{figure}

\subsection{Air Relative Velocity}
Since aerial vehicles generate forces within an air-relative medium, the corresponding constraints are established accordingly. The ground speed $\bm{V}^{\mathcal{G}}$ is determined by a wind triangle depicted as shown in~\reffig{fig:dubins_airplane} and can be expressed as
\begin{align}
    \bm{V}^{\mathcal{G}} = \bm{V}^{\mathcal{A}} + \bm{W}.
    \label{eq:wind_triangle}
\end{align}
Here, $\bm{V}^{\mathcal{A}}=[V^{\mathcal{A}}_x, V^{\mathcal{A}}_y, V^{\mathcal{A}}_z]^T$ represents the air-relative velocity of the aircraft. Considering fixed-wing vehicles' optimal range-efficiency at a specific airspeed, we assume a constant airspeed magnitude $\Bar{V}^\mathcal{A} = \|\bm{V}^\mathcal{A}\|$ for the vehicle. We define the air-relative flight path angle and heading:
\begin{align}
    \theta^{\mathcal{A}} &= \arctan2(V^{\mathcal{A}}_y, V^{\mathcal{A}}_x) &\quad \theta^{\mathcal{A}} &\in [0, 2\pi]\\
    \gamma^{\mathcal{A}} &= \arcsin(V^{\mathcal{A}}_z/\Bar{V}^\mathcal{A})&\quad \gamma^{\mathcal{A}} &\in [\gamma^\mathcal{A}_{min}, \gamma^\mathcal{A}_{max}].\nonumber
\end{align}

It should be noted that the maneuverability limit, especially the flight path angle, of the aircraft is inherently expressed in an air-relative context by design.

\section{APPROACH} \label{sec:approach}
\subsection{Ground-Relative Dubins Airplane Model in Wind}

In this work, we express the vehicle motion as a Dubins airplane path relative to the ground. In the first step, we compute the path according to \refequ{eq:dubins_airplane}. Then we compute the air-relative velocity $\bm{V}^{\mathcal{A}}$ such that the resulting ground velocity $\bm{V}^{\mathcal{G}}$ aligns with the path direction, denoted by the unit vector $\bm{u}_{\eta}$. We decompose \refequ{eq:wind_triangle} into its tangential component $()_\parallel$ and its normal component $()_\perp$ with respect to the path:

\begin{align}
    {V}^\mathcal{G}{\bm{u}}_{\eta} = & \bm{V}^\mathcal{A}_\parallel + \bm{W}_\parallel,\\
    0 =& \bm{V}^\mathcal{A}_\perp + \bm{W}_\perp.
    \label{eq:normal_equilibrium}
\end{align}

With the constant airspeed magnitude $\Bar{V}^\mathcal{A}$ and \refequ{eq:normal_equilibrium} we can compute the tangential airspeed component $V^\mathcal{A}_{\parallel}$ and ground speed:

\begin{align}
    V^\mathcal{A}_{\parallel} &= \pm\sqrt{(\Bar{V}^\mathcal{A})^2 - ({W}_{\perp}})^2,
    \label{eq:airspeed}
    \\
    {V}^\mathcal{G} &= {V}^\mathcal{A}_{\parallel} + {W}_{\parallel}.
\end{align}




Given that there are two solutions for ground speed $V^\mathcal{G}$, we select the larger positive one to maximize the progress of the path.

We identify two conditions under which a path is considered infeasible. First, if at any point along the path, the airspeed $\Bar{V}^\mathcal{A}$ falls below the perpendicular wind component ${W}_{\perp}$, the path is considered infeasible. This indicates the absence of a real solution for the parallel component of the velocity $V^\mathcal{A}_{\parallel}$, rendering the planned path unattainable regardless of the vehicle's heading. Secondly, if $V^\mathcal{G}$ becomes negative, it suggests that the desired direction of the path is unreachable, and thus infeasible given the wind condition and fixed airspeed constraint.

\subsection{Fixed-Wing Energy Model}
\label{sec:energy_model}
In this section, we introduce the energy model used in the planner. We assume a small angle of attack such that thrust $T$ aligns with airspeed. Furthermore, we assume constant drag $D$, due to the constant airspeed and small angle of attack assumption. The variation in drag is considerably less significant compared to the variation in lift for different flight regimes~\cite{Spera_2008}. Under these assumptions, the aircraft's dynamics can be simplified to a static longitudinal equilibrium:
\begin{align}
0 = T - D - mg \cdot \sin(\gamma^\mathcal{A}),
\end{align}
where $g$ is the gravitational constant, and $m$ represents the constant vehicle mass. Thus, the required thrust of the vehicle can be computed as a function of flight path angle $\gamma^\mathcal{A}$:
\begin{align}
    T \approx \max(D + mg sin(\gamma^\mathcal{A}), 0).
\end{align}
For descent, the required thrust may be less than zero indicating that the aircraft is accelerating. Since we assume a static equilibrium, we have a conservative energy estimate due to ignoring potential energy savings caused by such overspeed assuming the energy is dumped, for example by the use of spoilers. 

We define the energy rate model that maps the path into the energy consumed per unit of time as power $P:\mathcal{X}\mapsto\mathbb{R}$:
\begin{align}
    P = P_c + \frac{T \Bar{V}^\mathcal{A}}{c_T},
    \label{eq:energy_model}
\end{align}
where $P_c$ is the constant power consumed by the \ac{sUAV} avionics, and $c_T$ is the constant power coefficient of the thrust. However, this can be replaced by any arbitrary sophisticated energy model.


\subsection{Feasibility}
\label{sec:feasibility}
The feasibility of the path is checked by analyzing if the kinematic relations of each part can hold with the given wind and whether the path is in collision with the ground.

There are mainly two cases where the path becomes infeasible due to wind. First, the path is infeasible when the ground speed ${V}^\mathcal{G}$ calculated from~\refequ{eq:airspeed} has no solution, due to large wind speeds. Second, if the wind-relative flight path angle exceeds the limit, the path is considered infeasible. Both the vertical and horizontal wind components can influence the air-relative flight path angle. For example, the air-relative flight path angle will result in a steeper glide slope when traversing downwind, compared to traversing upwind. Thus, in the ground-relative limits $\gamma^\mathcal{G}_{min,max}$ should more set to smaller magnitudes than $\gamma^\mathcal{A}_{min,max}$ consider these cases.

Therefore, we can define the feasible space due to wind as the following:
\begin{align}
    \mathcal{W}^{+} = \{\bm{x}|\  {V}^\mathcal{G} > 0  \  \cap  \ \gamma^\mathcal{A}\in[\gamma^\mathcal{A}_{min}, \gamma^\mathcal{A}_{max}]\}.
    \label{eq:feasibility}
\end{align}

In case the vehicle is in collision with the terrain, the vehicle path intersects with the forbidden space $\mathcal{X}^{-}$.

The resulting safe set $\mathcal{X}_{safe}$ is defined as the intersection of the permitted space considering wind $\mathcal{W}^{+}$ and terrain $\mathcal{X}^{+}$
\begin{align}
    \mathcal{X}_{safe} = \mathcal{W}^{+} \cap \mathcal{X}^{+}.
\end{align}

\section{ENERGY-OPTIMIZED PLANNING}
In this section, we explain how the energy model introduced in~\refsec{sec:approach} is incorporated into the sampling-based planning framework\footnote{\href{https://github.com/ethz-asl/fw_planning}{https://github.com/ethz-asl/fw\_planning}}.
\subsection{Planner}

We use the sampling-based planner \ac{RRT*}~\cite{rrt*}. While the method is compatible with any other sampling-based planning method, we use \ac{RRT*} as it is asymptotically-optimal and probabilistically complete, thus rapidly finding a solution and refining it within the remaining planning time.

We plan in the Dubins airplane space~\cite{chitsaz2007time}, which connects states with ground relative Dubins airplane paths. We utilize the Dubins airplane path implementation of~\cite{mclain2014implementing} and take the suboptimal medium altitude case to speed up the computation~\cite{oettershagen2017towards}. Also, the Dubins set classification method~\cite{shkel2001classification} with corrections~\cite{lim2023circling} is used to speed up the computation of the Dubins path. 

\subsection{Cost Objective}

We use Euler forward integration along fixed-length segments with the discretized length $\Delta l$ to compute the cost according to \refequ{eq:energy_model}. As the wind is assumed to be constant within one integration step, the choice of $\Delta l$ and the integration scheme is a trade-off between computational cost and accuracy. The time to traverse a segment can be computed with the known groundspeed $\Delta t_i = \Delta l/{V}^\mathcal{G}_i$. Subsequently, the cost of a motion, or so to say, the energy consumed for the path is computed as:

\begin{align}
    E =& \begin{cases}
        \sum_{i=0} P(\bm{x_i}) \Delta t_i & \text{if $\bm{x_i} \in \mathcal{W}^+$}\\
        \infty & \text{if $\bm{x_i} \not\in \mathcal{W}^+$}
    \end{cases}
\end{align}
where we set the cost to infinity if there is any infeasible wind along the motion.

\subsection{Feasibility}
To guarantee that the paths are collision-free, states and motions are first checked using the motion validator with an elevation map of the terrain. Given a 2.5D elevation map (a 2D map with height), collision checking is performed by comparing the altitude of the state to the terrain elevation. Then, to verify whether the wind-related constraints are satisfied, we compute the path cost. Only collision-free motions with finite cost are considered feasible paths. This way, the wind field only needs to be accessed once, compared to accessing the wind field twice for feasibility and path cost computation.

\section{EVALUATION SETUP}
In this section, we explain the setup used for evaluating the proposed method.
\subsection{Setup}

We implement the energy-optimized planner using the \ac{OMPL}~\cite{sucan2012the-open-motion-planning-library} integrated into \ac{ROS} according to \cite{oettershagen2017towards} with a modified cost and motion model. All evaluations were run on an Intel Core i7-8565U CPU, which is comparable to the compute power that is available on real systems~\cite{lim2024safe}. 100 runs were executed for each planner configuration and scenario.

We used synthetic and realistic wind fields to evaluate the proposed planner compared to baseline methods. The synthetic wind fields on an empty map were hand-generated to highlight the differences between different planning methods. We generated the realistic wind fields using the \ac{WRF} model~\cite{skamarock2019description}. We utilize five nested domains with the coarsest domain at  \SI{9}{\kilo\meter} driven by the hourly ERA5 data to compute the winds with a 3:1 scaling ratio between the nests at the innermost domain at \SI{111}{\meter} resolution\footnote{\href{https://github.com/ethz-asl/wrf-sim}{https://github.com/ethz-asl/wrf-sim}}.

In this evaluation, we model the vehicle as \iac{sUAV}, such as the MakeflyEasy Believer, equipped with an onboard computer. The vehicle parameters used for the evaluation are presented in Table~\ref{tab:vehicle_parameters}. To accommodate a tailwind equal to the airspeed we set the climb angle limits using the following equation: $2 \gamma^{\mathcal{G}}_{max} = \gamma^{\mathcal{A}}_{max}$. We set the planning time for the synthetic wind fields to \SI{30}{\second} and increase it to \SI{60}{\second} for realistic wind fields to account for their much larger extent.
\begin{table}[t]
    \vspace{2mm}
    \centering
    \caption{Parameters of the fixed-wing vehicle model used for evaluation}
    \begin{tabular}{c c c c c c c c}
        \toprule
        $m$ & $\kappa_{max}$ & $c_T$ &  $\Bar{V}^{\mathcal{A}}$& $\gamma^{\mathcal{G}}_{max}$& $\gamma^{\mathcal{A}}_{max}$ & $D$ & $P_c$\\
        \midrule
        \SI{5}{\kilogram} & $0.02$ & 0.3 & \SI{15}{\metre/\second} & $10.0^\circ$ & $20.0^\circ$ & \SI{5}{\newton} & \SI{60}{\watt}\\
        \bottomrule
    \end{tabular}
    \label{tab:vehicle_parameters}
    \vspace{-4mm}
\end{table}

\begin{table*}[t]
\centering
\vspace{2mm}
\caption{The performance metrics of the different cost objectives and path formulations for the two synthetic wind fields. The shortest path planning (d\textsubscript{g}) is compared to time-optimized planning with air-relative paths (t\textsubscript{a}) and ground-relative (t\textsubscript{g}) and energy-optimized planning with the two path types (e\textsubscript{a} and e\textsubscript{g}).}
\small
\newcolumntype{C}{ @{}>{${}}c<{{}$}@{} }
\begin{tabular}{l l l *5{rCl}}
\toprule
 & & & \multicolumn{3}{c}{d\textsubscript{g}} & \multicolumn{3}{c}{t\textsubscript{a}} & \multicolumn{3}{c}{t\textsubscript{g}} & \multicolumn{3}{c}{e\textsubscript{a}} & \multicolumn{3}{c}{e\textsubscript{g}} \\

\toprule
\parbox[t]{2mm}{\multirow{3}{*}{\rotatebox[origin=c]{90}{Shear}}} & \parbox[t]{2mm}{\multirow{3}{*}{\rotatebox[origin=c]{90}{2 m/s}}} & Graph states [-] & \textbf{2383} &\pm & \textbf{205} & 575 &\pm & 10 & 562 &\pm & 18 & 513 &\pm & 4 & 487 &\pm & 23  \\
    & & Time [s]   & 235 &\pm & 1 & \textbf{234} &\pm & \textbf{1} & 235 &\pm & 1 & 236 &\pm & 1 & 236 &\pm & 2 \\
    & & Energy [kJ] & 73.9 &\pm & 0.2 & \textbf{72.6} &\pm & \textbf{0.3} & 73.0 &\pm & 0.5 & 73.0 &\pm & 0.5 & 73.3 &\pm & 0.5 \\
\midrule
    \parbox[t]{2mm}{\multirow{3}{*}{\rotatebox[origin=c]{90}{Shear}}} & \parbox[t]{2mm}{\multirow{3}{*}{\rotatebox[origin=c]{90}{5 m/s}}} & Graph states  [-] & \textbf{2462} &\pm & \textbf{62} & 476 &\pm & 12 & 522 &\pm & 22 & 423 &\pm & 11 & 492 &\pm & 13 \\
    & & Time [s]   & 305 &\pm & 1 & \textbf{239} &\pm & \textbf{2} & 244 &\pm & 3 & 243 &\pm & 4 & 247 &\pm & 3 \\
    & & Energy [J] & 94.5 &\pm & 0.3 & \textbf{74.5} &\pm & \textbf{1.0} & 76.2 &\pm & 1.4 & 75.2 &\pm & 1.0 & 76.5 &\pm & 1.0 \\
\midrule

    \parbox[t]{2mm}{\multirow{3}{*}{\rotatebox[origin=c]{90}{Shear}}} & \parbox[t]{2mm}{\multirow{3}{*}{\rotatebox[origin=c]{90}{10 m/s}}} & Graph states  [-] & \textbf{2450} &\pm & \textbf{156} & 388 &\pm & 6 & 514 &\pm & 18 & 359 &\pm & 6 & 498 &\pm & 16 \\
    & & Time [s]   & 609 &\pm & 2 & \textbf{234} &\pm & \textbf{3} & 243 &\pm & 5 & 236 &\pm & 4 & 245 &\pm & 5 \\
    & & Energy [J] & 188.7 &\pm & 0.6 & \textbf{72.4} &\pm & \textbf{1.0} & 75.9 &\pm & 2.0 & 73.2 &\pm & 1.3 & 75.9 &\pm & 1.6 \\
\midrule

    \parbox[t]{2mm}{\multirow{3}{*}{\rotatebox[origin=c]{90}{Shear}}} & \parbox[t]{2mm}{\multirow{3}{*}{\rotatebox[origin=c]{90}{15 m/s}}} & Graph states  [-] & \textbf{2453} &\pm & \textbf{330} & 247 &\pm & 11 & 804 &\pm & 31 & 232 &\pm & 5 & 767 &\pm & 18 \\
    & & Time [s]   & inf &\pm & inf & \textbf{228} &\pm & \textbf{4} & 235 &\pm & 8 & 230 &\pm & 6 & 236 &\pm & 8 \\
    & & Energy [J] & inf &\pm & inf & \textbf{70.6} &\pm & \textbf{1.3} & 74.6 &\pm & 3.2 & 71.3 &\pm & 1.8 & 73.3 &\pm & 2.5 \\
\toprule
\parbox[t]{2mm}{\multirow{3}{*}{\rotatebox[origin=c]{90}{Updraft}}} & \parbox[t]{2mm}{\multirow{3}{*}{\rotatebox[origin=c]{90}{5 m/s}}} & Graph states & \textbf{2245} & \mathbf{\pm} & \textbf{176} & 497 & \pm & 58 & 366 & \pm & 14 & 472 & \pm & 58 & 417 & \pm & 13 \\
    & & Time [s]    & 347 & \pm & 1 & \textbf{245} & \pm & \textbf{15} & 347 & \pm & 1  & 572 & \pm & 119 & 683 & \pm & 59 \\ 
    & & Energy [kJ] & 256 & \pm & 1 & 153 & \pm & 11 & 251 & \pm & 17 & 72 & \pm & 12   & \textbf{51} & \pm & \textbf{8} \\
\bottomrule
\end{tabular}
\label{tab:quantitative_results_synthetic}
\vspace{-2mm}
\end{table*}

\subsection{Baseline Methods}
We compare the energy-optimized planning to the shortest-distance~\cite{lim2024safe} and time-optimized planning with ground-relative Dubins airplane paths. Additionally, we also consider the air-relative Dubins airplane paths explored in previous work~\cite{oettershagen2017towards}, with time or energy used as the cost objective. For the reader's convenience, we present the air-relative Dubins model considering the air-relative properties and the wind in the equation of motion:


\textbf{\begin{align}
    f(\bm{\eta}(s), \bm{W}) = \begin{pmatrix}\cos(\gamma^{\mathcal{A}})\cos(\theta^\mathcal{A})+ \frac{W_x}{\Bar{V}^\mathcal{A}}\\
         \cos(\gamma^{\mathcal{A}})\sin(\theta^\mathcal{A})+ \frac{W_y}{\Bar{V}^\mathcal{A}} \\
         \sin(\gamma^{\mathcal{A}}) + \frac{W_z}{\Bar{V}^\mathcal{A}}\\
         cos(\gamma^{\mathcal{A}}) \kappa
         \end{pmatrix}.
    \label{eq:air_relative_dubins_airplane}
\end{align}}

There is no closed-form solution for the air-relative path connecting two states in a non-uniform wind field. Thus, we follow the iterative approach to compute such a path, as outlined in~\cite{oettershagen2017towards}. However, once the path is determined, all air-relative variables remain constant for the entire path. Therefore, the energy rate model shown in \refequ{eq:energy_model} can be solved by multiplying the air-relative distance, unlike the incremental integration of the ground-relative paths.

\section{SYNTHETIC WIND FIELD EXPERIMENTS} \label{experiments}

\subsection{Setup}

\begin{figure}[t]
\begin{subfigure}[b]{\columnwidth}
  \centering{\includegraphics[width=0.95\linewidth]{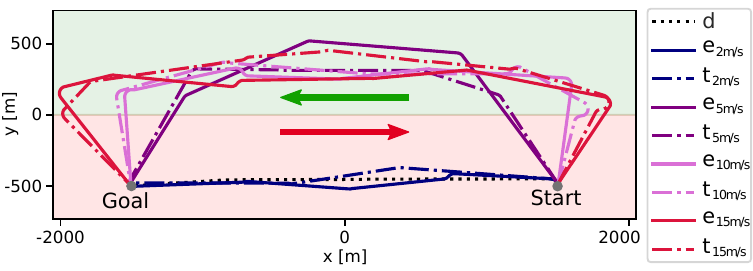}}
  \caption{\emph{Horizontal Shear} environment}
  \label{fig:qualitative_horizontal}
\end{subfigure}%

\begin{subfigure}[b]{\columnwidth}
  \centering{\includegraphics[width=0.95\linewidth]{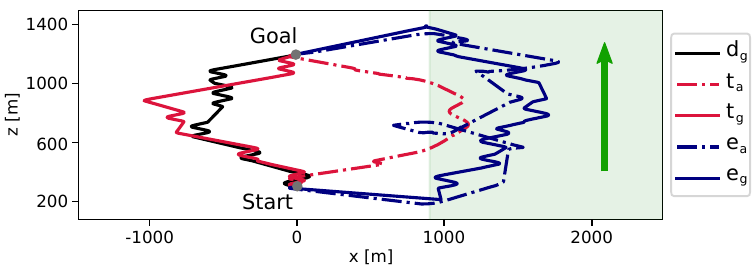}}
  \caption{\emph{Updraft} environment}
  \label{fig:qualitative_vertical}
\end{subfigure}%
\caption{Energy-optimized paths in a) \emph{Horizontal Shear} environment, b) \emph{Updraft} environment, explain the different paths d for shortest path, e for energy, t for time, subscript for wind field magnitude}
\label{fig:qualitative_paths_synthetic}
\vspace{-3mm}
\end{figure}

We evaluate our approach in synthetic wind fields to study the interaction between the path representation and optimization objective in non-uniform wind fields. We investigate two distinct environments, labeled as \emph{Horizontal Shear} and \emph{Updraft} shown in~\reffig{fig:qualitative_paths_synthetic}. The \emph{Horizontal Shear} environment features a wind field consisting solely of horizontal winds, with two regions where winds blow in opposite directions. Different wind magnitudes are investigated in this wind field as shown in \reftab{tab:quantitative_results_synthetic}. The \emph{Updraft} requires the airplane to gain altitude and contains an updraft region with \SI{5}{\metre/\second} at \SI{900}{\meter} distance from the start and goal position. Outside of the updraft region colored in green, there is no wind.

\ifshowfigure
\begin{figure*}
\vspace{2mm}
\centerline{\includegraphics[width=\linewidth]{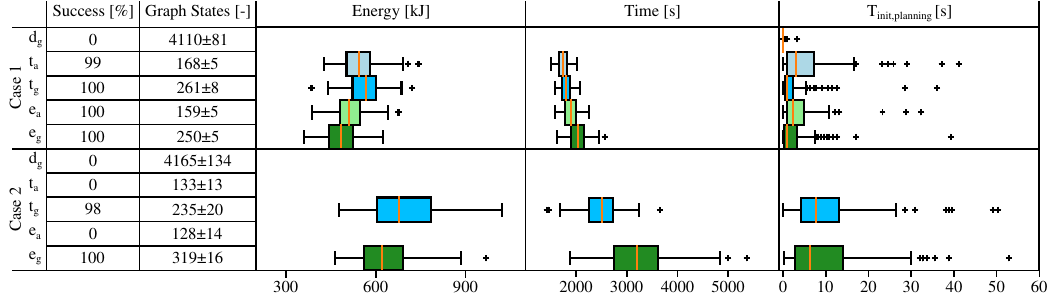}}
\caption{The performance metrics of the different cost objectives and path formulations for the realistic wind fields. The shortest path planning (d\textsubscript{g}) is compared to time-optimized planning with air-relative paths (t\textsubscript{a}) and ground-relative (t\textsubscript{g}) and energy-optimized planning with the two path types (e\textsubscript{a} and e\textsubscript{g}).}
\label{fig:quantitative_results_real}
\vspace{-3mm}
\end{figure*}
\fi

\subsection{Results}
The ground-relative paths planned with the different cost objectives for different wind speeds in \reffig{fig:qualitative_horizontal} within the \emph{Horizontal Shear} environment show that the wind-aware cost formulations result in detours to take advantage of the tailwinds when the winds are high enough. In scenarios with only horizontal wind, the energy- and time-optimized paths are equivalent due to the constant airspeed assumption.

Quantitative results can be found in~\reftab{tab:quantitative_results_synthetic}. Air-relative paths result in slightly lower costs compared to the ground-relative paths but are comparable in this environment. In the \emph{Updraft} environment, for both path types, the energy-optimized paths require less energy at the downside of a longer flight time as the path maximizes time spent in the updraft. Time-optimized paths demand less flight time but at the expense of more energy, as the aircraft is consistently climbing. In contrast to the energy-optimized planning, there is a large difference between the path formulations in the time-optimized planning. The ground-relative climb angle is bound to $\gamma^{\mathcal{G}}$, while the air-relative path with the wind support in the updraft can achieve higher climb angles, as evident in \reffig{fig:qualitative_vertical} explaining this difference.

A path planned with a wind-aware cost formulation is guaranteed to be safe subject to the wind field used during planning, while the shortest geometric path might be infeasible due to strong winds, as evidenced in the \SI{15}{\metre/\sec} case for the \emph{Horizontal Shear} environment.

Lastly, it is notable that the ground-relative planning in the \emph{Horizontal Shear} environment yields larger motion trees compared to air-relative methods. This effect is more pronounced for higher wind magnitudes. In the \emph{Updraft} environment, we observe the opposite trend with slightly larger trees for the air-relative methods. This occurs because the paths are in general long in this scenario, significantly slowing down the cost computation of ground-relative paths due to the numerical integration. Overall, wind-aware planning is significantly slower than shortest-path planning.



\ifshowtable

\begin{table*}[t]
\centering
\caption{The performance metrics of the different cost objectives and path formulations for the realistic wind fields. The shortest path planning (d\textsubscript{g}) is compared to time-optimized planning with air-relative paths (t\textsubscript{a}) and ground-relative (t\textsubscript{g}), as well as energy-optimized planning with the two path types (e\textsubscript{a} and e\textsubscript{g}).}
\small
\newcolumntype{C}{ @{}>{${}}c<{{}$}@{} }
\begin{tabular}{l l *5{rCl}}
\toprule
 & & \multicolumn{3}{c}{d\textsubscript{g}} & \multicolumn{3}{c}{t\textsubscript{a}} & \multicolumn{3}{c}{t\textsubscript{g}} & \multicolumn{3}{c}{e\textsubscript{a}} & \multicolumn{3}{c}{e\textsubscript{g}} \\

\toprule
    \parbox[t]{2mm}{\multirow{5}{*}{\rotatebox[origin=c]{90}{Case 1}}} & Graph states [-] & \textbf{4110} &\pm & \textbf{81} & 168 &\pm & 5 & 261 &\pm & 8 & 159 &\pm & 5 & 250 &\pm & 5  \\
    & Time [s]   & inf &\pm & inf & \textbf{1755} &\pm & \textbf{111} & 1809 &\pm & 108 & 1895 &\pm & 142 & 2030 &\pm & 176 \\
    & Energy [J] & inf &\pm & inf & 547 &\pm & 62 & 561 &\pm & 58 & 517 &\pm & 50 & \textbf{488} &\pm & \textbf{54} \\
    & T\textsubscript{first solution} [s] & \textbf{0.06} &\pm & \textbf{0.34} & 5.7 &\pm & 7.6* & 2.5 &\pm & 5.0 & 4.1 &\pm & 5.9 & 2.8 &\pm & 5.0 \\
    & Success [\%] & \multicolumn{3}{c}{0} & \multicolumn{3}{c}{99} & \multicolumn{3}{c}{\textbf{100}} & \multicolumn{3}{c}{\textbf{100}} & \multicolumn{3}{c}{\textbf{100}} \\
\midrule
    \parbox[t]{2mm}{\multirow{5}{*}{\rotatebox[origin=c]{90}{Case 2}}} & Graph states [-] & \textbf{4165} &\pm & \textbf{134} & 133 &\pm & 13 & 325 &\pm & 20 & 128 &\pm & 14 & 319 &\pm & 16  \\
    & Time [s]   & inf &\pm & inf & inf &\pm & inf & \textbf{2555} &\pm & \textbf{787} & inf &\pm & inf & 3310 &\pm & 1156 \\
    & Energy [J] & inf &\pm & inf & inf &\pm & inf & 728 &\pm & 290 & inf &\pm & inf & \textbf{650} &\pm & \textbf{178} \\
    & T\textsubscript{first solution} [s] & \textbf{0.007} &\pm & \textbf{0.01} & \multicolumn{3}{c}{-} & 10.6 &\pm & 10.2 & \multicolumn{3}{c}{-} & 10.2 &\pm & 10.5 \\
    & Success [\%] & \multicolumn{3}{c}{0} & \multicolumn{3}{c}{0} & \multicolumn{3}{c}{98} & \multicolumn{3}{c}{0} & \multicolumn{3}{c}{\textbf{100}} \\

\bottomrule
\end{tabular}
\label{tab:quantitative_results_real}
\vspace{-0mm}
\end{table*}
\fi




\section{Realistic wind field experiments}
\subsection{Setup}

\begin{figure}
\centerline{\includegraphics[width=\linewidth]{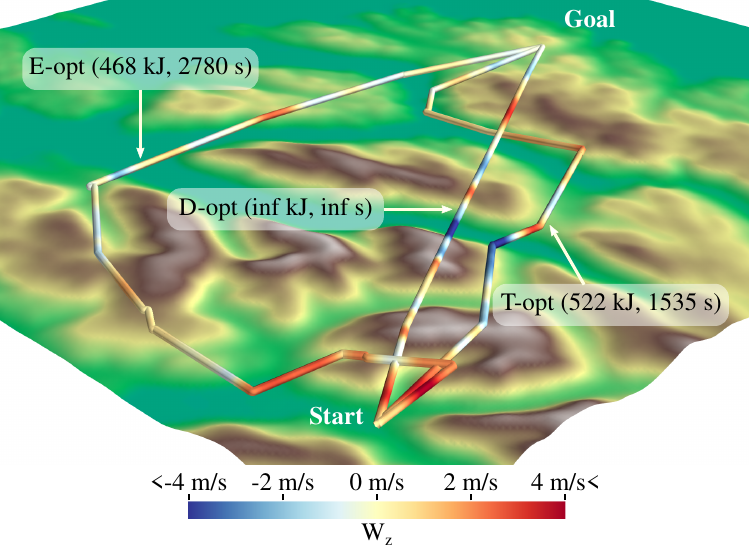}}
\caption{The planned paths for case 2 with WRF generated winds colored by the encountered vertical wind. The energy-optimized path (E-opt) leverages the updrafts and avoids strong downdrafts resulting in a longer path compared to the time-optimized path (T-opt) that goes through a strong downdraft. The shortest path (D-opt) does not consider any wind information and is infeasible to track due to the strong downdraft around the middle.}
\label{fig:real_case2}
\vspace{-3mm}
\end{figure}

To evaluate the proposed method, we evaluate the planner's performance in realistic wind fields computed by \ac{WRF}. Experiments are conducted on two different terrains with an extent of $\SI{30}{\kilo\meter}\times\SI{30}{\kilo\meter}$ each. The first case, illustrated in \reffig{fig:real_case1}, is centered around the \SI{1798}{\meter} high mountain, Rigi, located in central Switzerland. The second case is situated around the hills of the Isle of Bute in Scotland, as depicted in \reffig{fig:real_case2}. We compare our planning method with the same baselines as those in the synthetic wind field experiments.

\subsection{Results}
In these two cases, we observe that the paths computed with the different cost objectives vary significantly. The shortest-distance paths result in infinite costs for both cases as strong encountered downdrafts exceeding the \ac{sUAV}'s limitation do not allow safely following these paths. The time-optimized paths leverage mostly the horizontal wind and allow flying through strong downdrafts if they are within the limitations. The energy-optimized paths leverage existing updrafts to reduce energy compared to the time optimal paths by \qtyrange{11}{13}{\percent} albeit at a trade-off of \qtyrange{12}{30}{\percent} longer flight times as shown in \reffig{fig:quantitative_results_real}.

In these complex non-uniform wind fields computing the air-relative paths requires more computational time compared to the synthetic wind fields that still had large uniform regions. Thus, the planner takes longer to find good solutions as evidenced by the number of graph states and the time to find the first solution. The large variance of the time to first solution is a result of the stochasticity of the underlying RRT* planner. In the second case, the air-relative variants failed to find an initial solution in all runs in contrast to the ground-relative variants where only \SI{2}{\percent} failed for time-optimized planning.




\section{DISCUSSION}
In this work, we explored different cost formulations and path representations. The shortest-path planning is least computationally demanding but in realistic settings can result in unsafe paths. The energy-optimized planning leverages the wind to plan more energy-efficient paths at the trade-off for longer flight time compared to time-optimized paths. However, these paths are only different in the presence of vertical wind. The solutions with the air-relative Dubins paths tend to be shorter/more efficient than the ground-relative counterparts. This is because in the air-relative path formulation the \ac{sUAV} "drifts with the wind" whereas the latter treats wind as a disturbance. This can be seen in \reffig{fig:qualitative_vertical} where within the updraft the air-relative paths fully leverage the upwind achieving a high climb rate, while the ground-relative formulation inherently fights the updraft to stay on the pre-planned geometric path achieving much smaller climb rates even for the time-optimized case.

Planning with air-relative paths is not suitable for planning onboard with an \ac{sUAV}, as the iterative computation of these paths in non-uniform wind fields is too computationally expensive. Using the ground-relative path formulation yields a much faster planner that can consistently find solutions within a reasonable time frame. Furthermore, air-relative paths are less suited for path following due to the challenges of predicting the air-relative paths with the large uncertainties associated with wind sensing.

\section{CONCLUSIONS} \label{con}
In this work, we proposed an energy-optimized planner for non-uniform wind fields using geometric Dubins airplane paths. We evaluated our approach in a series of experiments, comparing it to the baseline cost and path formulations, and discussed the trade-offs between different formulations. We expect this approach to enable more efficient operations for fixed-wing vehicles covering long distances.

This work can be extended in several ways. Firstly, handling variable airspeed allows for more efficient paths, as the energy-optimal cruise speed depends on the encountered winds. Secondly, while we in this work assume a temporally constant wind field, extending the approach to cope with the  uncertainty in predicted wind fields enables planning for more robust paths in case of wind prediction errors. Additionally, to address changing weather conditions, future work could explore how to incorporate temporally varying wind fields into the planning framework.

\addtolength{\textheight}{0cm}   





\bibliographystyle{IEEEtran}
\bibliography{references}

\end{document}

%% file: include/acronyms.tex
\DeclareAcronym{sUAS}{
  short = sUAS,
  long  = small uncrewed aerial system,
  short-indefinite = an,
  long-indefinite = a
}

\DeclareAcronym{sUAV}{
  short = sUAV,
  long  = small uncrewed aerial vehicle,
  short-indefinite = an,
  long-indefinite = a
}

\DeclareAcronym{RoC}{
  short = RoC,
  long  = rate of climb,
  short-indefinite = an,
  long-indefinite = a
}

\DeclareAcronym{DEM}{
  short = DEM,
  long  = digital elevation map
}

\DeclareAcronym{OMPL}{
  short = OMPL,
  long  = Open Motion Planning Library,
  short-plural =  ,
  long-plural = 
}

\DeclareAcronym{ROS}{
  short = ROS,
  long  = Robot Operating System,
  short-indefinite = a
}

\DeclareAcronym{RRT*}{
  short = RRT*,
  long  = Rapidly-exploring Random Tree Star,
  short-plural =  ,
  long-plural = 
}

\DeclareAcronym{CFD}{
  short = CFD,
  long  = Computational Fluid Dynamics,
  short-plural =  ,
  long-plural = 
}

\DeclareAcronym{WRF}{
    short = WRF,
    long = weather research \& forecasting model,
    short-indefinite = a,
    long-indefinite = a
}

%% file: include/newcommands.tex
\newcommand{\reffig}[1]{Fig.~\ref{#1}}
\newcommand{\reftab}[1]{Table~\ref{#1}}
\newcommand{\refsec}[1]{Section~\ref{#1}}

\newcommand{\refequ}[1]{Eq.~\eqref{#1}}

%% file: main.bbl
\begin{thebibliography}{10}
\providecommand{\url}[1]{#1}
\csname url@samestyle\endcsname
\providecommand{\newblock}{\relax}
\providecommand{\bibinfo}[2]{#2}
\providecommand{\BIBentrySTDinterwordspacing}{\spaceskip=0pt\relax}
\providecommand{\BIBentryALTinterwordstretchfactor}{4}
\providecommand{\BIBentryALTinterwordspacing}{\spaceskip=\fontdimen2\font plus
\BIBentryALTinterwordstretchfactor\fontdimen3\font minus \fontdimen4\font\relax}
\providecommand{\BIBforeignlanguage}[2]{{%
\expandafter\ifx\csname l@#1\endcsname\relax
\typeout{** WARNING: IEEEtran.bst: No hyphenation pattern has been}%
\typeout{** loaded for the language `#1'. Using the pattern for}%
\typeout{** the default language instead.}%
\else
\language=\csname l@#1\endcsname
\fi
#2}}
\providecommand{\BIBdecl}{\relax}
\BIBdecl

\bibitem{COLOMINA2014unmanned}
I.~Colomina and P.~Molina, ``Unmanned aerial systems for photogrammetry and remote sensing: A review,'' \emph{ISPRS Journal of Photogrammetry and Remote Sensing}, vol.~92, pp. 79--97, 2014.

\bibitem{ezequiel2014uav}
C.~A. Ezequiel, M.~Cua, N.~C. Libatique, G.~L. Tangonan, R.~Alampay, R.~T. Labuguen, C.~M. Favila, J.~L. Honrado, V.~Canos, C.~Devaney, and et~al., ``Uav aerial imaging applications for post-disaster assessment, environmental management and infrastructure development,'' \emph{2014 International Conference on Unmanned Aircraft Systems (ICUAS)}, 2014.

\bibitem{schopferer2015performance}
S.~Schopferer and T.~Pfeifer, ``Performance-aware flight path planning for unmanned aircraft in uniform wind fields,'' in \emph{2015 International Conference on Unmanned Aircraft Systems (ICUAS)}.\hskip 1em plus 0.5em minus 0.4em\relax IEEE, 2015, pp. 1138--1147.

\bibitem{stastny2019flying}
T.~Stastny and R.~Siegwart, ``On flying backwards: Preventing run-away of small, low-speed, fixed-wing uavs in strong winds,'' in \emph{2019 IEEE/RSJ International Conference on Intelligent Robots and Systems (IROS)}.\hskip 1em plus 0.5em minus 0.4em\relax IEEE, 2019, pp. 5198--5205.

\bibitem{boon2017comparison}
M.~A. Boon, A.~P. Drijfhout, and S.~Tesfamichael, ``Comparison of a fixed-wing and multi-rotor uav for environmental mapping applications: A case study,'' \emph{The International Archives of the Photogrammetry, Remote Sensing and Spatial Information Sciences}, vol.~42, pp. 47--54, 2017.

\bibitem{ch_windatlas}
T.~Schlegel, M.~Geissmann, M.~Hertach, and D.~Kr\"{o}pfli, ``Windatlas {Schweiz}: Jahresmittel der modellierten windgeschwindigkeit und windrichtung,'' Federal Department of Environment, Transport, Energy and Communications, Tech. Rep. COO.2207.110.2.1073455, 2016.

\bibitem{lim2024safe}
J.~Lim, F.~Achermann, R.~Girod, N.~Lawrance, and R.~Siegwart, ``Safe low-altitude navigation in steep terrain with fixed-wing aerial vehicles,'' \emph{IEEE Robotics and Automation Letters}, vol.~9, no.~5, pp. 4599--4606, 2024.

\bibitem{moon2023time}
B.~Moon, S.~Sachdev, J.~Yuan, and S.~Scherer, ``Time-optimal path planning in a constant wind for uncrewed aerial vehicles using dubins set classification,'' \emph{IEEE Robotics and Automation Letters}, 2023.

\bibitem{techy2009minimum}
L.~Techy and C.~A. Woolsey, ``Minimum-time path planning for unmanned aerial vehicles in steady uniform winds,'' \emph{Journal of guidance, control, and dynamics}, vol.~32, no.~6, pp. 1736--1746, 2009.

\bibitem{bucher2023robust}
T.~Bucher, T.~Stastny, S.~Verling, and R.~Siegwart, ``Robust wind-aware path optimization onboard small fixed-wing uavs,'' in \emph{AIAA SCITECH 2023 Forum}, 2023, p. 2640.

\bibitem{chung2015learning}
J.~J. Chung, N.~R. Lawrance, and S.~Sukkarieh, ``Learning to soar: Resource-constrained exploration in reinforcement learning,'' \emph{The International Journal of Robotics Research}, vol.~34, no.~2, pp. 158--172, 2015.

\bibitem{chakrabarty2013uav}
A.~Chakrabarty and J.~Langelaan, ``Uav flight path planning in time varying complex wind-fields,'' in \emph{2013 American control conference}.\hskip 1em plus 0.5em minus 0.4em\relax IEEE, 2013, pp. 2568--2574.

\bibitem{wirth2015}
L.~Wirth, P.~Oettershagen, J.~Amb\"{u}hl, and R.~Siegwart, ``Meteorological path planning using dynamic programming for a solar-powered {UAV},'' in \emph{2015 IEEE Aerospace Conference}, March 2015, pp. 1--11.

\bibitem{oettershagen2017towards}
P.~Oettershagen, F.~Achermann, B.~M{\"u}ller, D.~Schneider, and R.~Siegwart, ``Towards fully environment-aware uavs: Real-time path planning with online 3d wind field prediction in complex terrain,'' \emph{arXiv preprint arXiv:1712.03608}, 2017.

\bibitem{chitsaz2007time}
H.~Chitsaz and S.~M. LaValle, ``Time-optimal paths for a dubins airplane,'' in \emph{2007 46th IEEE conference on decision and control}.\hskip 1em plus 0.5em minus 0.4em\relax IEEE, 2007, pp. 2379--2384.

\bibitem{skamarock2019wrf}
W.~C. Skamarock, J.~B. Klemp, J.~Dudhia, D.~O. Gill, Z.~Liu, J.~Berner, W.~Wang, J.~G. Powers, M.~G. Duda, D.~M. Barker \emph{et~al.}, ``A description of the advanced research wrf model version 4,'' \emph{National Center for Atmospheric Research: Boulder, CO, USA}, vol. 145, no. 145, p. 550, 2019.

\bibitem{bry2015aggressive}
A.~Bry, C.~Richter, A.~Bachrach, and N.~Roy, ``Aggressive flight of fixed-wing and quadrotor aircraft in dense indoor environments,'' \emph{The International Journal of Robotics Research}, vol.~34, no.~7, pp. 969--1002, 2015.

\bibitem{dubins1957curves}
L.~E. Dubins, ``On curves of minimal length with a constraint on average curvature, and with prescribed initial and terminal positions and tangents,'' \emph{American Journal of mathematics}, vol.~79, no.~3, pp. 497--516, 1957.

\bibitem{mclain2014implementing}
M.~Owen, R.~W. Beard, and T.~W. McLain, ``Implementing dubins airplane paths on fixed-wing uavs*,'' \emph{Handbook of Unmanned Aerial Vehicles}, p. 1677–1701, Aug 2014.

\bibitem{seemann2014rrt}
M.~Seemann and K.~Janschek, ``Rrt-based trajectory planning for fixed wing uavs using bezier curves,'' in \emph{ISR/Robotik 2014; 41st International Symposium on Robotics}.\hskip 1em plus 0.5em minus 0.4em\relax VDE, 2014, pp. 1--8.

\bibitem{lee2014optimal}
D.~Lee, H.~Song, and D.~H. Shim, ``Optimal path planning based on spline-rrt for fixed-wing uavs operating in three-dimensional environments,'' in \emph{2014 14th International conference on control, automation and systems (ICCAS 2014)}.\hskip 1em plus 0.5em minus 0.4em\relax IEEE, 2014, pp. 835--839.

\bibitem{achermann2024windseer}
F.~Achermann, T.~Stastny, B.~Danciu, A.~Kolobov, J.~J. Chung, R.~Siegwart, and N.~Lawrance, ``Windseer: real-time volumetric wind prediction over complex terrain aboard a small uncrewed aerial vehicle,'' \emph{Nature Communications}, vol.~15, no.~1, p. 3507, 2024.

\bibitem{patrikar2020wind}
J.~Patrikar, B.~G. Moon, and S.~Scherer, ``Wind and the city: Utilizing uav-based in-situ measurements for estimating urban wind fields,'' in \emph{2020 IEEE/RSJ International Conference on Intelligent Robots and Systems (IROS)}.\hskip 1em plus 0.5em minus 0.4em\relax IEEE, 2020, pp. 1254--1260.

\bibitem{Spera_2008}
\BIBentryALTinterwordspacing
D.~A. Spera, ``Models of lift and drag coefficients of stalled and unstalled airfoils in wind turbines and wind tunnels - nasa technical reports server (ntrs),'' Oct 2008. [Online]. Available: \url{https://ntrs.nasa.gov/search.jsp?R=20090001311}
\BIBentrySTDinterwordspacing

\bibitem{rrt*}
S.~Karaman and E.~Frazzoli, ``Incremental sampling-based algorithms for optimal motion planning,'' \emph{Robotics: Science and Systems VI}, 2010.

\bibitem{shkel2001classification}
A.~M. Shkel and V.~Lumelsky, ``Classification of the dubins set,'' \emph{Robotics and Autonomous Systems}, vol.~34, no.~4, pp. 179--202, 2001.

\bibitem{lim2023circling}
J.~Lim, F.~Achermann, R.~B{\"a}hnemann, N.~Lawrance, and R.~Siegwart, ``Circling back: Dubins set classification revisited,'' in \emph{Workshop on Energy Efficient Aerial Robotic Systems, International Conference on Robotics and Automation 2023}, 2023.

\bibitem{sucan2012the-open-motion-planning-library}
I.~A. {\c{S}}ucan, M.~Moll, and L.~E. Kavraki, ``The {O}pen {M}otion {P}lanning {L}ibrary,'' \emph{{IEEE} Robotics \& Automation Magazine}, vol.~19, no.~4, pp. 72--82, December 2012, \url{https://ompl.kavrakilab.org}.

\bibitem{skamarock2019description}
W.~C. Skamarock, J.~B. Klemp, J.~Dudhia, D.~O. Gill, Z.~Liu, J.~Berner, W.~Wang, J.~G. Powers, M.~G. Duda, D.~M. Barker \emph{et~al.}, ``A description of the advanced research wrf version 4,'' \emph{NCAR tech. note ncar/tn-556+ str}, vol. 145, 2019.

\end{thebibliography}
